%% file: main.tex
\documentclass[letterpaper, 10 pt, conference]{ieeeconf}
\IEEEoverridecommandlockouts
                                                          
\usepackage{amsmath,amssymb,amsfonts}
\usepackage{graphicx}
\usepackage{textcomp}
\usepackage{xcolor}
\usepackage{hyperref}
\usepackage{amsmath}
\usepackage[noend]{algpseudocode}
\usepackage{comment}
\usepackage[normalem]{ulem}
\usepackage[colorinlistoftodos]{todonotes}
\usepackage[framemethod=tikz]{mdframed}
\newtheorem{dfn}{Definition}
\newtheorem{asm}{Assumption}

\newtheorem{problem}{Problem}

\def\BibTeX{{\rm B\kern-.05em{\sc i\kern-.025em b}\kern-.08em
    T\kern-.1667em\lower.7ex\hbox{E}\kern-.125emX}}

\usepackage{graphicx}
\usepackage{psfrag}
\usepackage{pstool}
\usepackage{multirow}
\usepackage{soul}
\algnewcommand\algorithmicforeach{\textbf{for each}}
\algdef{S}[FOR]{ForEach}[1]{\algorithmicforeach\ #1\ \algorithmicdo}

\DeclareUnicodeCharacter{0301}{\'{e}}

\usepackage{xspace}

\makeatletter

\DeclareRobustCommand\onedot{\futurelet\@let@token\@onedot}
\def\@onedot{\ifx\@let@token.\else.\null\fi\xspace}

\usepackage{algorithm,algpseudocode,setspace}

\usepackage{etoolbox}
\makeatletter
\patchcmd{\@makecaption}
  {\scshape}
  {}
  {}
  {}
\makeatletter
\patchcmd{\@makecaption}
  {\\}
  {.\ }
  {}
  {}
\makeatother
\def\tablename{Table}

\input{bmit2e}

\input{mymacro}

\usepackage[style=ieee,doi=false,mincitenames=1,maxcitenames=20,minbibnames=1,maxbibnames=20,isbn=false,url=true]{biblatex}
\makeatother
\usepackage[belowskip=-7pt,aboveskip=10pt,font={small}]{caption}
\usepackage[font={small}]{subcaption}

\makeatletter

\begin{document}

\title{ 
Augment-Connect-Explore: a Paradigm for  Visual Action \\Planning with Data Scarcity 
}

\author{Martina Lippi*$^{1}$, Michael C. Welle*$^{2}$, Petra Poklukar$^{2}$, Alessandro Marino$^{3}$, Danica Kragic$^{2}$
\thanks{*These authors contributed equally (listed in alphabetical order).}
\thanks{ ${}^1$Roma Tre University, Rome, Italy  {\tt\small martina.lippi@uniroma3.it} }%
\thanks{ ${}^2$KTH Royal Institute of Technology Stockholm, Sweden, {\tt\small \{mwelle, poklukar, dani\}@kth.se}}%
\thanks{ ${}^3$University of Cassino and Southern Lazio, Cassino, Italy {\tt\small al.marino@unicas.it}}%
\thanks{This work was supported by the Swedish Research Council, Knut and Alice Wallenberg Foundationm, by the European Research Council (ERC-884807), by the European Commission (Project CANOPIES-101016906), and  by Dipartimento di Eccellenza granted to DIEI
Department, University of Cassino and Southern Lazio.}
}

\maketitle

\begin{abstract}
Visual action planning particularly excels in applications where the state of the system cannot be computed explicitly, such as manipulation of deformable objects, as it enables planning directly from raw images. 
Even though the field has been significantly accelerated by deep learning techniques, a crucial requirement for their success is the availability of a large amount of data.
In this work, we propose the Augment-Connect-Explore (ACE) paradigm to enable visual action planning in cases of data scarcity. 
 We build upon the Latent Space Roadmap (LSR) framework which performs planning with a graph built in a low dimensional latent space. In particular, ACE is used to \emph{i)} Augment the available training dataset by autonomously creating new pairs of datapoints, \emph{ii)} create new unobserved Connections among representations of states in the latent graph, and \emph{iii)} Explore new regions of the latent space in a targeted manner. We validate the proposed approach on both simulated  box stacking and real-world folding task showing the applicability for rigid and deformable object manipulation tasks, respectively. 
\end{abstract}

\input{includes/intro-rw.tex}

\input{includes/problem.tex}

\input{includes/method.tex}

\input{includes/experiments.tex}

\input{includes/conclusions.tex}


\AtNextBibliography{\scriptsize}

\printbibliography

\end{document}

%% file: bmit2e.tex
\font\bfmath=cmmib10
\mathchardef\Gamma="7100
\mathchardef\Delta="7101
\mathchardef\Theta="7102
\mathchardef\Lambda="7103
\mathchardef\Xi="7104
\mathchardef\Pi="7105
\mathchardef\Sigma="7106
\mathchardef\Upsilon="7107
\mathchardef\Phi="7108
\mathchardef\Psi="7109
\mathchardef\Omega="710A

\mathchardef\alpha="710B
\mathchardef\beta="710C
\mathchardef\gamma="710D
\mathchardef\delta="710E
\mathchardef\epsilon="710F
\mathchardef\zeta="7110
\mathchardef\eta="7111
\mathchardef\theta="7112
\mathchardef\iota="7113
\mathchardef\kappa="7114
\mathchardef\lambda="7115
\mathchardef\mu="7116
\mathchardef\nu="7117
\mathchardef\xi="7118
\mathchardef\pi="7119
\mathchardef\rho="711A
\mathchardef\sigma="711B
\mathchardef\tau="711C
\mathchardef\upsilon="711D
\mathchardef\phi="711E
\mathchardef\chi="711F
\mathchardef\psi="7120
\mathchardef\omega="7121
\mathchardef\epsilon="7122

\mathchardef\varepsilon="7122
\mathchardef\vartheta="7123
\mathchardef\varpi="7124
\mathchardef\varrho="7125
\mathchardef\varsigma="7126
\mathchardef\varphi="7127
\mathchardef\imath="717B
\mathchardef\jmath="717C

%% file: mymacro.tex
\def\smallbfW{{\raise1.5pt\hbox{\mbox{\boldmath $_W$}}}}

\def\red{\color{red}}

\def\blue{\color{blue}}

\def\my4psfrag#1#2#3#4#5#6#7#8{
        \begin{figure}[htp]
        \begin{center}
            \begin{tabular}[h]{c c}
              {\leavevmode{\includegraphics[width=#1truecm]{#2.eps}}}
              &
              {\leavevmode{\includegraphics[width=#1truecm]{#3.eps}}} \\
              {\leavevmode{\includegraphics[width=#1truecm]{#4.eps}}}
              &
              {\leavevmode{\includegraphics[width=#1truecm]{#5.eps}}}
         \end{tabular}
           \vspace{#6}
           \caption{#7}
           \label{#8}
        \end{center}
        \end{figure}
}

\def\mydouble4psfrag#1#2#3#4#5#6#7#8{
        \begin{figure*}[htp]
        \begin{center}
            \begin{tabular}[h]{c c}
              {\leavevmode{\includegraphics[width=#1truecm]{#2.eps}}}
              &
              {\leavevmode{\includegraphics[width=#1truecm]{#3.eps}}} \\
              {\leavevmode{\includegraphics[width=#1truecm]{#4.eps}}}
              &
              {\leavevmode{\includegraphics[width=#1truecm]{#5.eps}}}
         \end{tabular}
           \vspace{#6}
           \caption{#7}
           \label{#8}
        \end{center}
        \end{figure*}
}

%% file: includes/intro-rw.tex
\section{Introduction}
\label{sec:intro}

Given a start \emph{observation} of the system, the goal of visual action planning~\cite{wang2019learning} 
is to produce \textit{i)} an action plan comprised of the actions required to reach a desired state, and \textit{ii)} a visual plan containing observations, i.e., images, of
intermediate states that  will be traversed during the execution of the planned actions. In this way, the planner can be given raw image observations.  This aspect is crucial in applications where the state of the system cannot be easily described analytically,   as for instance  in the manipulation of deformable objects like wires in manufacturing settings, clothes in fashion industries or food in  agricultural setups  as in the European Project CANOPIES. 
The supporting visual plan additionally improves the interpretability of these methods~\cite{wang2019learning,pmlr_liu20h}. 

\begin{figure}[t]
    \centering
    \includegraphics[width=0.9\linewidth]{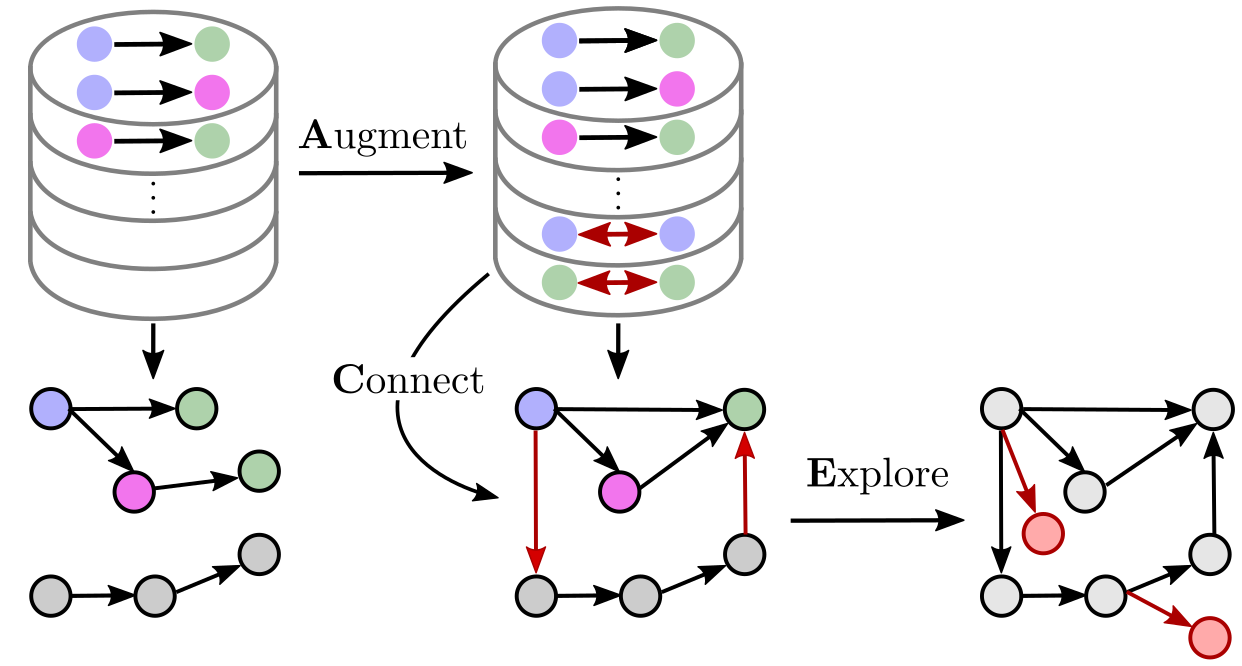}
    \caption{Overview of our ACE paradigm: (1) gaining new similar datapairs by \textbf{A}ugmenting existing ones, (2) building unseen \textbf{C}onnections in the latent space, and (3) efficiently \textbf{E}xploring new regions. Color represent the underlying states of the system (see Sec.~\ref{sec:proplemdef} for details).}
    \label{fig:overview}
\end{figure}

While it has been shown that visual dynamics used for planning can be learned directly from images,  several approaches considered planning in low-dimensional latent space (discussed in Sec.~\ref{sec:rw}).  These methods reduce the complexity of planning in the image space but generally  depend on vast amount of data to train reliable policies as they require long rollouts for successful planning. In practice, this can hinder their applicability to real robotic hardware. 

Therefore, in this work, we propose a method for performing visual action planning in case of data scarcity. We  build upon our Latent Space Roadmap (LSR) framework \cite{iros,bestlsr} that  learns a low-dimensional latent space from input images and  builds a graph, in this latent space,  that is used to perform planning. 
We tackle data scarcity  by introducing the \textit{Augment-Connect-Explore} (ACE) paradigm that is based on:  
\textit{i) Augment:} creating new informative pairs of datapoints  exploiting demonstrated actions  to improve the latent space structure.
\textit{ii) Connect:} increasing the connectivity of the LSR by building new connections, i.e.  \textit{shortcuts}, among nodes to improve the capability to traverse the latent space. 
\textit{iii) Explore:} proposing \emph{targeted} exploratory actions by leveraging latent representations as well as collected actions to explore new states in an efficient manner. 
Our contributions can be summarized as follows:
\begin{itemize}
    \item We introduce the ACE paradigm to address data scarcity in visual action planning by augmenting, connecting and exploring. 
    \item  To realize ACE, we design a novel Suggestion Module
    that proposes possible actions from a given state. This is used in conjunction with a simple neural network that predicts the next latent state given the current state and desired action.
    \item We thoroughly analyze the individual and cumulative effects  of ACE components on a simulated box stacking task and  demonstrate improved performance of the combined ACE framework on a real-world folding task  under data scarcity.
\end{itemize}

\section{Related Work} \label{sec:rw}
Several approaches learn the visual dynamics directly from images and use it for planning. 
In~\cite{wang2019learning} visual foresight plans for deforming a rope into desired configurations are generated with Context Conditional Causal InfoGANs.
The learned rope inverse dynamics is then considered to reach the configurations in the generated plan. In~\cite{finn2017deep} Long-Short Term Memory blocks are used to compose a video prediction model predicting the stochastic pixel flow from frame to frame given the action. This model is then integrated in a Model Predictive Control (MPC) framework to produce  visual plans and push objects of interest. 
Building on the visual foresight frameworks,  the work in \cite{fabric_vsf_2020} 
proposes the VisuoSpatial Foresight which integrates  the depth map information with the pure RGB data to learn the visual dynamic model of fabrics in a simulated environment. An extension of this approach is given  in \cite{Daniel_AR2021} where the main  steps of the framework are improved. 

To reduce the complexity of planning in the image space, low-dimensional latent spaces  have been explored in several studies, e.g., \cite{Ichter2019}-\cite{savinov2018semiparametric}. A framework for global search in latent space is presented in  \cite{Ichter2019}, where motion planning is performed directly in this latent space using an RRT-based algorithm with collision checking and latent space dynamics modelled as neural networks.  Contrastive learning is used in~\cite{Yan2020LearningPR} to derive a predictive model in the latent space that is exploited to find rope and cloth flattening actions.
Latent space goal-conditioned predictors, formulated as hierarchical models,  are introduced~\cite{pertch2020long} to limit the search space  to trajectories  that lead to the goal configuration and thus to perform long-horizon visual planning. Latent planning has also been successfully applied in Reinforcement Learning (RL) settings, like for example in \cite{rafailov2021offline} for model-based offline RL and in \cite{pmlr-v80-haarnoja18a} for hierarchical RL. The combination of RL with graph structures in the latent  space is explored in~\cite{savinov2018semiparametric}, where a node is created for  each encoded  observation.  Building on~\cite{savinov2018semiparametric}, temporal closeness of the consecutive observations in the trajectories is also exploited in~\cite{pmlr_liu20h}.  However, the above methods generally require a large amount of long rollouts for successful planning. 
Therefore, in this work, we tackle visual action planning for scenarios with scarce training data.

%% file: includes/problem.tex
\section{Preliminaries and Problem Statement} \label{sec:proplemdef}
 In this section we provide preliminary notions 
 for our framework and formalize the problem of visual action planning  with data scarcity. 

\subsection{Dataset structure}\label{sec:dataset}
\begin{figure}
    \centering
    \includegraphics[width=0.8\linewidth]{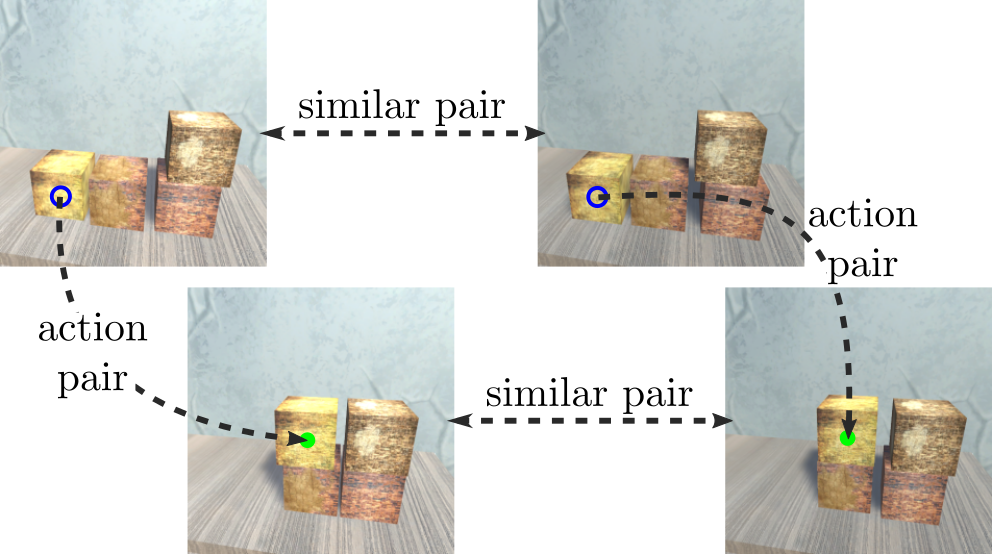}
    \caption{Examples of similar and action pairs. The similar pairs show the same underlying state, while the observations in the action pairs show different underlying states. }
    \label{fig:sim_dissim}
    \vspace{-\baselineskip}
\end{figure}

Let $\mathcal{O}$ be the space of all possible observations, i.e., images, of the system states. 
We consider a training dataset $\mathcal{T}_o$ consisting of $q$ tuples  $(O_1, O_2, \rho)$, where $O_1$ is an observation of the start state, $O_2$ an observation of the successor state, and $\rho$ a variable denoting the respective action between the states.  Here, an action is defined as a single transformation that brings the system to a new state different from the starting one. For example, in Fig.~\ref{fig:sim_dissim}, an action corresponds to moving a box. The variable $\rho = (a, u)$ is composed of a binary variable $a \in \{0, 1\}$ indicating whether or not an action occurred and a variable $u$ containing the task-dependent action-specific information in case an action occurred, i.e. $a=1$. 
We say that no action was performed, i.e., \mbox{$a = 0$}, if observations $O_1$ and $O_2$ are different variations of the same (unknown) underlying state of the system.   In the bottom row of Fig.~\ref{fig:sim_dissim}, the observations exhibit  lightning and slight positional  variations, but correspond to the same underlying state of the system determined by the arrangement of the boxes. We refer to  a tuple in the form $(O_1, O_2, \rho = (1, u))$ as an \textit{action pair} and  $(O_1, O_2, a = 0)$ as a  \textit{similar pair} (shown in Fig.~\ref{fig:sim_dissim}). 

\subsection{Visual Action Planning}
Let $\mathcal{U}$ be the set of possible  actions of the system.   
%
A visual action plan is the combination of an action plan $P_u$ and a visual plan $P_o$ that  lead the system from a given start $O_s\in\mathcal{O}$ to a goal observation $O_g\in\mathcal{O}$, i.e., such that $P_o = \{O_s = O_1, O_2,...,O_N=O_g\}$  and $P_u = \{u_1, u_1,...,u_{N-1}\}$, where $u_n \in \mathcal{U}$ produces a transition between consecutive observations $O_n$ and $O_{n+1}$ for each $n\in\{0, ..., N-1\}$. 

To reduce the complexity of the problem, we build a low-dimensional latent space $\mathcal{Z}$ encoding $\mathcal{O}$ that aims to capture the  underlying states of the system.
 
\begin{dfn}\label{defn::map}
The \emph{latent mapping} function $\xi: \mathcal{O} \to \mathcal{Z}$ maps an observation $O_n\in\mathcal{O}$ into its latent representation  $z_n\in\mathcal{Z}$. The \emph{observation generator} function $\omega: \mathcal{Z} \to \mathcal{O}$ retrieves a possible observation $O_n\in\mathcal{O}$ associated with a latent representation $z_n\in\mathcal{Z}$.
\end{dfn}

\begin{dfn}\label{defn::latentdynamic}
The \emph{latent dynamic function}  \mbox{$f: \mathcal{Z} \times \mathcal{U} \to \mathcal{Z}$} transitions the system through the latent space.
\end{dfn}

Given these functions, one way to realize visual action planning is  to map the start and goal observations in the latent space, obtaining $z_s = \xi(O_s)$, $z_g = \xi(O_g)$, and then perform planning directly in $\mathcal{Z}$   by exploiting the latent dynamic function $f$.  
This leads to the definition of an action plan $P_u$ with corresponding latent plan $P_z = \{z_{s} = z_1, z_2, ..., z_N = z_{g}\}$, based on which the visual plan $P_o$ is generated through the observation generator $\omega$.  
Note that in practice the functions $\xi$, $\omega$ and $f$ are  unknown and need to be approximated. The quality of these approximations depends on the available amount of observations $\mathcal{T}_o$ of the system.




\subsection{Problem Statement}

When $\mathcal{T}_o$ is \emph{scarce}, it might not contain all possible latent states associated with the system as well as possible transitions among them. In this work we are interested in solving the following problem. 

\begin{problem}\label{prob:}
Given a \emph{scarce} training dataset $\mathcal{T}_o$ as well as a start $O_s\in\mathcal{O}$ and a goal observation $O_g\in\mathcal{O}$, our objective is to find the related visual action plan $(P_o, P_u)$. 
\end{problem}

To solve it, we exploit a simple insight that two latent states are similar if  the same set of actions can be applied to both, and if these, in turn, also yield the same set of consecutive states. 
We first define the set of actions that can be applied to a given latent state $z$.


\begin{dfn} \label{defn::sugg}
A \emph{suggestion function}  $\eta: \mathcal{O} \to  \mathcal{U}$ provides a subset $\eta(O_i) = \mathcal{U}_i \subseteq\mathcal{U}$ of actions that can be applied from an observation $O_i \in \mathcal{O}$.
\end{dfn}

Given the suggestion function $\eta$ and latent dynamic function $f$,  we define similar states as follows,   in line with bisimulation theory \cite{GIVAN2003163}.

\begin{dfn}\label{con::sm}
The states $z_i, z_j \in  \mathcal{Z}$ corresponding to the observations $O_i, O_j \in \mathcal{O}$ are said to be \emph{similar}  if
\begin{itemize}
    \item $\eta(O_i) = \eta(O_j) = \widetilde{\mathcal{U}}$, and
    \item $\{ f(z_i, u) \} = \{ f(z_j, u)\}$ for every $u \in \widetilde{\mathcal{U}}$.
\end{itemize}
\end{dfn}

\subsection{Latent Space Roadmap Framework}\label{sec:lsr}
We addressed the problem of visual action planning for scenarios of \emph{complete} training datasets, i.e., those covering all possible states and transitions among them, in our earlier works \cite{iros, bestlsr} by introducing the LSR framework,  which we briefly recall in the following. The basic idea of this framework is to perform planning in the low dimensional latent space $\mathcal{Z}$ by \emph{i)} structuring it  to respect the underlying states of the system, and \emph{ii)}  building a graph directly in this latent space to guide the planning. 

To address point \emph{i)}, we define the concept of covered regions. 
We map the training dataset $\mathcal{T}_o$ described in Section~\ref{sec:dataset} into the latent space $\mathcal{Z}$ to obtain
 a set of \emph{covered} states $\mathcal{T}_z = \{z_1, ..., z_{2 q}\} \subset \mathcal{Z}$, i.e., \mbox{$ \mathcal{T}_z = \xi(\mathcal{T}_o)$}, for which we make the following assumption as in \cite{iros,bestlsr}. 
\begin{asm}
\label{asm::eps-validity}
Given a  covered state $z \in \mathcal{T}_{z}$, there exists $\varepsilon > 0$ such that any other state $z'$ in the $\varepsilon-$neighborhood $N_{\varepsilon}(z)$ of $z$ can be considered as the same underlying state. 
\end{asm} 

We define the  union of $\varepsilon$-neighbourhoods of the covered states $z \in \mathcal{T}_z$ as \emph{covered subspace} 
\begin{equation}
\label{eq:union-eps} \mathcal{Z}_{sys} = \bigcup\nolimits_{z \in \mathcal{T}_z} N_{\varepsilon}(z) \subset \mathcal{Z},
\end{equation} 
which can be rewritten as the union of $m$ path-connected components \cite{bestlsr} called \emph{covered regions} and denoted by $\{\mathcal{Z}_{sys}^i\}_{i=1}^{m}$.
Note that in a well structured latent space, each covered region encodes a possible underlying state of the system.
We  define a set of transitions that connect covered regions. 
A \emph{covered transition function}  \mbox{$f^{ij}_{sys}: \mathcal{Z}^{i}_{sys} \times \mathcal{U} \to \mathcal{Z}^{j}_{sys}$} maps a point $z^i \in \mathcal{Z}^{i}_{sys}$ to a point $z^j \in \mathcal{Z}^{j}_{sys}$ when applying an action $u\in\mathcal{U}$, with $i, j \in \{1, 2,..., m\}$ and $i \neq j$.
Given $\mathcal{Z}_{sys}$ and the covered transition functions $f^{ij}_{sys}$, we then define the Latent Space Roadmap:
\begin{dfn}
\label{def:lsr}
A Latent Space Roadmap is a directed graph $LSR = (\mathcal{V}_{LSR}, \mathcal{E}_{LSR})$ where each vertex \mbox{$v_i \in \mathcal{V}_{LSR} \!\subset\! \mathcal{Z}_{sys}$} for $i \in \{1, ..., m\}$ is a representative of the covered region $\mathcal{Z}^{i}_{sys}  \subset \mathcal{Z}_{sys}$, and each edge $e_{i, j} = (v_i, v_j) \in \mathcal{E}_{LSR}$ represents a covered transition function $f^{ij}_{sys}$ between the corresponding covered regions $\mathcal{Z}^{i}_{sys}$ and $\mathcal{Z}^{j}_{sys}$ for $i \neq j$. 
\end{dfn}



Two main modules compose the LSR framework. First, a Mapping Module (MM) implements the mapping function $\xi$ and observation generator $\omega$ defined in Def. \ref{defn::map} with a VAE framework. These are learned using a contrastive loss term, also called \emph{action} term, which attracts states belonging to similar pairs and repels states belonging to action pairs to a minimum distance $d_m$.  Second, an LSR module implements the LSR defined in Def. \ref{def:lsr} by applying clustering in $\mathcal{Z}$ to approximate the covered regions $\mathcal{Z}^{i}_{sys}$. Each obtained cluster is associated with a node in the LSR and edges among them are created using action pairs in the training dataset $\mathcal{T}_o$. In this process, average action specifics are also endowed in the edges to retrieve the action plan $P_u$ (see the Action Averaging Baseline in \cite{bestlsr} for details). 

%% file: includes/method.tex
\section{Overview of the approach}

\begin{figure}[t]
    \centering
    \includegraphics[width=\linewidth]{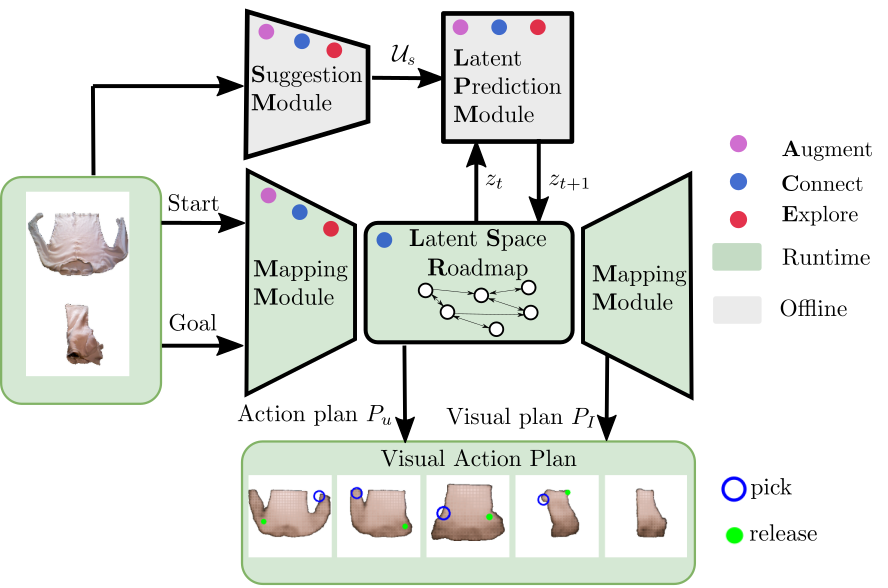}
    \caption{  Overview of the proposed architecture  for visual action planning. The modules involved in the ACE phases are highlighted with respective colored dots. The modules only used offline are marked in grey, while the ones also used online are marked in green. }
    \label{fig:overview}
    \vspace{-\baselineskip}
\end{figure}

In order to perform visual planning in case of data scarcity, the proposed ACE paradigm aims to:  
\begin{enumerate}
    \item  Augment the available dataset $\mathcal{T}_o$, autonomously creating new similar pairs. 
    \item Create new unobserved connections in the latent space $\mathcal{Z}$, increasing the set of  covered transition functions $f^{ij}_{sys}$ and number of respective edges in the LSR. 
    \item Explore new regions of $\mathcal{Z}$ in an efficient and guided manner, increasing the covered subspace $\mathcal{Z}_{sys}$. 
\end{enumerate}
To realise the ACE paradigm we extend the LSR framework with a Latent Prediction Model (LPM) and a Suggestion Module (SM). 
An overview of the overall ACE architecture 
including all the modules is shown in Fig. \ref{fig:overview}. We mark in green the modules used at run time to produce the visual action plan (bottom) from start $O_s$ and goal observations $O_g$ (left), and in grey the ones only used offline. 

The LPM module approximates the latent dynamics function in Def. \ref{defn::latentdynamic} which, given a latent state $z_i$ and an action $u \in \mathcal{U}$, predicts a potential next state  $z_{i+1}$. 
Note that LPM implicitly assumes a given MM.
The SM module approximates the suggestion function $\eta$ in Def. \ref{defn::sugg}  and, given an observation $O_i$,   suggests a set of potential actions $\mathcal{U}_i$ that are possible to perform. The input image  $O_i$ can be either an observation of the current state or an observation generated by $\omega$. 

To realize point 1), we rely on the definition of similar states in Def.~\ref{con::sm} and employ SM and LPM to find novel similar pairs  that are added to $\mathcal{T}_o$ to obtain the augmented training dataset $\overline{\mathcal{T}_o}$. 
The latter is then used to obtain a new mapping function approximation ${\xi}$ by  updating the MM, which leads to an enhanced structure of the latent space $\mathcal{Z}$. 

Regarding point 2), we use SM and LPM along with the covered subspace $\mathcal{Z}_{sys}$ to identify previously unseen transitions $f^{ij}_{sys}$ that are possible to execute, called \emph{valid} transitions. These are added to the LSR in the form of new edges referred to as \textit{shortcuts}.


 Similarly, point 3) is realized by using the 
SM, that suggests the set of possible actions $\mathcal{U}_i$ from the current state $z_i$, and the LPM that predicts potential next states. Among the possible actions, the most promising one is chosen for exploration, as described in Sec.~\ref{sec:exploration}.
This enables exploring new regions of the latent space $\mathcal{Z}$ in a guided manner. 

The ACE paradigm improves the individual components of the LSR framework which is then used to perform visual action planning. Note that even though we focus on the LSR framework, ACE is general and applicable to many other contexts, e.g., the proposed targeted exploration approach can be easily integrated into an RL setting.

\subsection{Models for LPM and SM}\label{sec:lpm-sm}
We model LPM as a simple multilayer perceptron (MLP), as detailed in Sec.~\ref{sec:exp}. During augmentation and connection phases, we also leverage the covered subspace $\mathcal{Z}_{sys}$ defined in~\eqref{eq:union-eps}: we consider a state $z_j$ predicted by the LPM reliable only if it falls within the covered subspace, i.e., $z_j\in\mathcal{Z}_{sys}$.  We refer to the LPM including the covered subspace check as reliable LPM (LPM-R) in the following.

The SM is built based on two core  considerations:  \emph{i)} 
 several  valid actions can be applied to the same state,   and \emph{ii)} the same action can be applied to different states. 
 We model the suggestion function $\eta$ with a Siamese network  trained with a contrastive loss that encourages clustering of the states from which the same subset of actions
can be performed. 
 
In detail, we build the training dataset for the SM, denoted by $\mathcal{T}^{SM}_o$, by rearranging the observations in the  training tuples in  $\mathcal{T}_o$ depending on the actions. A similar pair $(O_1,O_2, s = 1)$, where  $s$ is the similarity signal, is added to $\mathcal{T}^{SM}_o$ if the same action specifics $u$ is applied from $O_1$ and $O_2$ in $\mathcal{T}_o$, i.e., if there exist \mbox{$(O_1, - , \rho=(1,u))\in \mathcal{T}_o$} and \mbox{$(O_2, - , \rho=(1,u))\in \mathcal{T}_o$}, where $-$ denotes any other observation. 
On the other hand, a dissimilar pair $(O_1,O_2, s = 0)$ is added to $\mathcal{T}^{SM}_o$ when different action specifics are applied from $O_1$ and $O_2$ in $\mathcal{T}_o$, i.e., if there exist \mbox{$(O_1, - , \rho=(1,u_1))\in \mathcal{T}_o$} and \mbox{$(O_2, - , \rho=(1,u_2))\in \mathcal{T}_o$} with $u_1 \neq u_2$. Note that training the Siamese network with the dataset $\mathcal{T}^{SM}_o$  results in a latent space $\mathcal{Z}'$ different from $\mathcal{Z}$. 
The latent space $\mathcal{Z}'$ is then clustered and each cluster 
\mbox{$\mathcal{C}\subset \mathcal{Z}'$} is labeled with the set   $\mathcal{U}_c \subseteq \mathcal{U}$ containing all the actions  that are executed starting from the states points in $\mathcal{C}$. We experimentally validate that the above procedure allows to achieve a good approximation of the suggestion function  despite some possible erroneous similarity signals in $\overline{\mathcal{T}}_o^{SM}$. 

\begin{figure}
    \centering
    \includegraphics[width=0.65\linewidth]{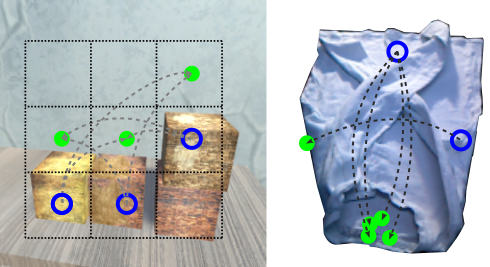}
    \vspace{-5pt}
    \caption{ Example of the set of suggested actions $\mathcal{U}$ using the SM for a box stacking (left) and T-shirt folding (right) task. The blue rings mark pick locations, while the green circles place locations. }
    \label{fig:sm_master}
\end{figure}

At run time, a novel observation $O_i$ is fed into the Siamese network to obtain its latent representation $z_i'\in\mathcal{Z}'$ and the set of suggested actions $\mathcal{U}_i$ associated with the closest cluster $\mathcal{C}_i$ as visualized in Fig. \ref{fig:sm_master}.

\section{ACE Paradigm}
In this section we present  the individual components of our ACE paradigm and provide an overview of the full framework. 

\subsection{Augment}
The proposed augmentation procedure builds new similar pairs  based on the definition in Def.~\ref{con::sm}. In particular, if the same set of actions applied from different observations $O_1$ and $O_2$ leads to the same underlying states, we consider the two starting observations as a similar pair. In doing so, we discover similar pairs among states that are  erroneously further apart in the latent space $\mathcal{Z}$. 
This occurs in practice since the latent mapping $\xi$ is only an approximation. Therefore, to improve the structure of $\mathcal{Z}$, it is crucial to identify more similar pairs in the dataset $\mathcal{T}_o$ such that ${\xi}$ is re-learned to map the same underlying states close together.

 Note that no labels about the underlying states contained in the training observations, that could be exploited for augmenting the dataset, are provided. In contrast, we only have access to the information of whether two observations are similar or there is an action between them.

\setlength{\textfloatsep}{10pt}

\begin{algorithm} \caption{Augmentation Algorithm}
\small
\def\negsp{\vspace{-5pt}}
\def\negup{\vspace{-7pt}}
\def\negin{\vspace{-3pt}}
\setstretch{1.2}
\begin{algorithmic}[section]
\Require Training dataset $\mathcal{T}_o$, search radius $r$
    \begin{algorithmic}[1]
    \State $\mathcal{T}_z \gets \text{MM}(\mathcal{T}_o)$ 
    \State $\overline{\mathcal{T}}_o := {\mathcal{T}}_o$
    \ForEach {$z_i \in \mathcal{T}_z $}
            \State $\mathcal{U}_i \gets \text{SM}(O_i)$
            \State $\mathcal{L}_i \gets \text{search in radius}(\mathcal{T}_z, z_i, r)$
            \State $\mathcal{L}_i \gets \text{descent  sort}(\mathcal{L}_i)$
            \State found := False
        \ForEach{$z_j\in \mathcal{L}_i$ and not found}
            \State  $O_j \gets \text{get observation}( \mathcal{T}_{o}, j)$ 
            \State $\mathcal{U}_j \gets \text{SM}(O_j)$
            \If{$\mathcal{U}_i \equiv \mathcal{U}_j$}
                \State $\mathcal{Z}^p_i, \mathcal{Z}^p_j \gets \text{LPM-R}(z_i, \mathcal{U}_i), \text{LPM-R}(z_j, \mathcal{U}_i)$
                \State $\mathcal{Z}^n_i, \mathcal{Z}^n_j \gets \text{nearest}(\mathcal{T}_z, \mathcal{Z}^p_i), \text{nearest}(\mathcal{T}_z, \mathcal{Z}^p_j)$
                    \If{ $\mathcal{Z}^n_i \equiv \mathcal{Z}^n_j$}
                        \State $\bar{\mathcal{T}}_o := \overline{\mathcal{T}}_o\cup \{\left(O_i, O_j, a = 0\right)\}$
                        \State found := True
                    \EndIf
            \EndIf
        \EndFor
    \EndFor
    \end{algorithmic}
\Return $\bar{\mathcal{T}}_o$
\end{algorithmic}
\label{alg::augment}

\end{algorithm}

Algorithm \ref{alg::augment} summarizes the augmentation procedure. Given the training dataset $\mathcal{T}_o$ and a search radius $r$ determining the search area around covered states, 
we encode all observations $O_i \in \mathcal{T}_o$ to obtain $\mathcal{T}_z \subset \mathcal{Z}$ (line~1) and initialize the augmented dataset $\overline{\mathcal{T}}_o$ (line~2). 
For each latent state $z_i \in \mathcal{T}_z$, we check if a new similar pair can be identified. We first obtain the set  $\mathcal{U}_i$ of possible actions from $z_i$ using the SM (line~4). Then, we define the set $\mathcal{L}_i$ of  covered latent states  which are within the search radius $r$ (line~5),  i.e., $\mathcal{L}_i = \{ z_j \in \mathcal{T}_z \,|\, z_j \in {N}_{r}(z_i) \}$. 
This is followed by a descent  sorting with respect to the distance  of each $z_j \in \mathcal{L}_i$ to $z_i$ (line~6). Note that we limit the search in a radius only for  computational reasons. Since the latent space $\mathcal{Z}$ already has a certain structure inferred from the non-augmented dataset $\mathcal{T}_o$  during training of MM, we avoid checking states that are too far away from the current and likely not similar to it.

At this point, we analyze the covered states $z_j\in\mathcal{L}_i$. Starting from the first $z_j$, we take the corresponding observation $O_j$ in the training dataset (line~9) and obtain the set of potential actions $\mathcal{U}_j$ (line~10).
If all the actions in the sets $\mathcal{U}_i,\,\mathcal{U}_j$ coincide, we  obtain the  sets of respective  predicted states  
$\mathcal{Z}^p_i, \mathcal{Z}^p_j$~(line~12) by the LPM-R, which checks the reliability condition discussed in Sec. \ref{sec:lpm-sm}. 
Based on these sets, the  sets $\mathcal{Z}^n_i, \mathcal{Z}^n_j$ consisting of closest covered latent states in $\mathcal{T}_z$ with respect to $\mathcal{Z}^p_i, \mathcal{Z}^p_j$, respectively, are found. If $\mathcal{Z}^n_i, \mathcal{Z}^n_j$ coincide, a new similar pair  $(O_i, O_j, a = 0)$ is added to the augmented dataset $\overline{\mathcal{T}}_o$, otherwise the next state $z_j\in\mathcal{L}_i$ is analyzed.

\subsection{Connect}
A good connectivity of nodes is essential for the success of  graph-based planning methods. Although more connections can be built by collecting more data, a more efficient approach involves building \textit{shortcuts}, i.e., connections between nodes  that are not directly induced by the training set.  In this work, we infer them by using the SM and LPM modules.
 Note that it is  crucial to add correct shortcuts as erroneous connections can be very detrimental for the graph planning capabilities, leading to unfeasible plans. 

\begin{algorithm} \caption{Connection Algorithm}
\small
\def\negsp{\vspace{-5pt}}
\def\negup{\vspace{-7pt}}
\def\negin{\vspace{-3pt}}
\setstretch{1.2}
\begin{algorithmic}[section]
\Require  ${LSR} = (\mathcal{V}_{LSR}, \mathcal{E}_{LSR})$, neighborhood size $\epsilon$
    \begin{algorithmic}[1]
    \ForEach {$z_i \in \mathcal{V}_{LSR} $}
        \State $O_i \gets \text{MM}(z_i)$
        \State $\mathcal{U}_i \gets \text{SM}(O_i)$
        \ForEach{ $ u \in \mathcal{U}_i$}
            \State $z_n \gets \text{LPM}(z_i, u)$
            \If{$\|z_j - z_n\|_1 < \varepsilon \text{ for } z_j \in \mathcal{V}_{LSR}, i \ne j$}
                 \State $\mathcal{E}_{{LSR}} \gets $ create edge ($z_i,z_j, u$)
            \EndIf
        \EndFor
    \EndFor
    \end{algorithmic}
\Return $LSR$
\end{algorithmic}
\label{alg::connect}
\end{algorithm}

Algorithm \ref{alg::connect} summarizes the proposed method for building shortcuts. The basic idea is that if an action $u$ suggested by the SM in a certain state $z_i$ leads to a covered state $z_j$, then the respective transition can be considered as valid and can be added to the LSR. 
In detail, given the LSR and the neighborhood size $\varepsilon$, 
we iterate over the states in the set of nodes $\mathcal{V}_{LSR}$ of the LSR. For each state $z_i$ in $\mathcal{V}_{LSR}$, we generate the respective observation $O_i$ through the observation generator $\omega$ of the MM (line~2) and obtain the set of potential actions $\mathcal{U}_i$ by the SM. For each $u \in \mathcal{U}_i$, LPM predicts the next state $z_n$ obtained from $z_i$ (line~5). If the predicted next state $z_n$ falls in the $\epsilon$-neighborhood of any other state $z_j \in \mathcal{V}_{LSR}$ in the LSR with $i\neq j$, an edge between $z_i$ and $z_j$ is added in the edge set $\mathcal{E}_{LSR}$ of the LSR (line~7). We also endow the edge with the new predicted action $u$ for action planning purposes as discussed in \ref{sec:lsr}.

\subsection{Explore}\label{sec:exploration}

The challenges of finding suitable  actions for exploration of the latent space $\mathcal{Z}$ 
are twofold: \emph{i)} finding valid actions that can be performed in the current state, and \emph{ii)} choosing the action that is most beneficial to the system. 

The SM model provides a solution to the first problem as it outputs a set of valid actions $\mathcal{U}_i$ for an observation $O_i$ corresponding to the current state $z_i$ as described in Sec.~\ref{sec:lpm-sm}. For the second problem, we propose to undertake the action that leads to the most unexplored area of the latent space $\mathcal{Z}$ at each exploration step.

The approach is summarized in Algorithm \ref{alg::explore}. Given the training dataset $\mathcal{T}_z$ and the  current state observation $O_i$, we first map both into $\mathcal{Z}$ with the mapping function $\xi$ of the MM (lines~1-2). Then, we retrieve the set of potential actions $\mathcal{U}_i$ from the current state $z_i$ through the SM. We initialize an empty auxiliary exploration list $\mathcal{L}_e$.   For each action $u \in \mathcal{U}_i$, we predict the next state $z_n$ using  the LPM  (line~6) and compute the distance $d_i$ from $z_i$ to the nearest covered state in $\mathcal{T}_z$ (line~7). The tuple given by the action $u$ and distance $d_i$ is added to the exploration list $\mathcal{L}_e$. 
Once all the actions in $\mathcal{U}_i$ have been analyzed, we return the action $u_e$ (line~9) that leads to the furthest latent state as the exploratory one. 
As described in the following, the observation $O_{i+1}$ obtained after executing $u_e$ is used to create a new action pair $(O_i, O_{i+1}, \rho= (1, u_e))$ that is added to $\mathcal{T}_o$. The latent space is explored by executing Algorithm~\ref{alg::explore} $n_e$ times.  Note that in case the LPM does not provide an accurate prediction for unseen states, this would only result in computing an imprecise  distance $d_i$, and then choosing a less beneficial action to explore. 

\begin{algorithm} \caption{Exploration Algorithm}
\small
\def\negsp{\vspace{-5pt}}
\def\negup{\vspace{-7pt}}
\def\negin{\vspace{-3pt}}
\setstretch{1.2}
\begin{algorithmic}[section]
\Require Training dataset $\mathcal{T}_o$, current observation $O_i$
    \begin{algorithmic}[1]
    \State $\mathcal{T}_z \gets \text{MM}(\mathcal{T}_o)$
    \State $z_i \gets \text{MM}(O_i)$
    \State $\mathcal{U}_i \gets \text{SM}(O_i)$
    \State $\mathcal{L}_e$ := $\{\}$
    \ForEach {$u \in \mathcal{U}_i$ }
        \State $z_n \gets \text{LPM}(z_i,u)$
        \State $d_i \gets \text{nearest}(\mathcal{T}_z, z_n)$
        \State $\mathcal{L}_e \gets \text{add tuple} (u, d_i)$ 
    \EndFor
    \State $u_e \gets \text{get action to furthest state} (\mathcal{L}_e)$
    \end{algorithmic}
\Return $u_e$
\end{algorithmic}
\label{alg::explore}
\end{algorithm}

\subsection{LSR with ACE}
In this section, we describe how the ACE components are combined within the LSR framework, summarized in 
Algorithm~\ref{alg::integration}.  Given  an initial training dataset $\mathcal{T}_o$, the hyperparameters $r$ and $\varepsilon$ as well as the number of total exploration steps $n_e$, we first build the models employed in the ACE paradigm (line 1). Secondly, we generate the augmented dataset $\overline{\mathcal{T}_o}$ following Algorithm \ref{alg::augment} with the search radius $r$ and use it to update the MM, LPM, and SM models. 
Thirdly, we perform the targeted exploration phase. For each exploration step $i\in\{1,...,n_e\}$, we get the current observation $O_i$, determine the most promising exploration action $u_e$ using 
Algorithm \ref{alg::explore} (line 6) and execute it (line 7) to reach the new observation $O_{i+1}$. The observed tuple $\left(O_i, O_{i+1}, (\rho=(1, u_e))\right)$ is added to the dataset $\overline{\mathcal{T}_o}$ (line 8). After  completing  the exploration, the LSR is built using the approach in \cite{iros} with neighborhood threshold $\varepsilon$ (line 9). Finally, we add the shortcuts as in Algorithm~\ref{alg::connect} (line 10) and the final LSR is returned. In case the system performance  after executing Algorithm~\ref{alg::integration} is not satisfactory, this can be repeated multiple times  to further improve the results.

\begin{algorithm} \caption{Integration Algorithm}
\small
\def\negsp{\vspace{-5pt}}
\def\negup{\vspace{-7pt}}
\def\negin{\vspace{-3pt}}
\setstretch{1.2}
\begin{algorithmic}[section]
\Require Training dataset $\mathcal{T}_o$, search radius $r$, neighborhood threshold $\varepsilon$,  number of explorations $n_e$
    \begin{algorithmic}[1]
    \State MM, LPM, SM $\gets$ build models$(\mathcal{T}_o)$
    \State ${\overline{\mathcal{T}}}_o \gets \text{augment dataset}(\mathcal{T}_o, r)$ [Alg.~\ref{alg::augment}]
    \State MM, LPM, SM $\gets$ update models$({\overline{\mathcal{T}}}_o)$
    \ForEach{$i\in\{1,..,n_{e}\}$}
        \State $O_i \gets \text{current observation} $
        \State $u_e \gets \text{get exploration action} ({\overline{\mathcal{T}_o}}, O_i)$ [Alg.~\ref{alg::explore}]
        \State $O_{i+1} \gets \text{perform action}(u_e)$
        \State ${\overline{\mathcal{T}_o}} := {\overline{\mathcal{T}_o}} \cup \{\left(O_i, O_{i+1}, (\rho=(1, u_e))\right)\}$
    \EndFor
    \State $LSR \gets \text{build LSR} ({\overline{\mathcal{T}_o}}, \varepsilon)$ \cite{iros}
    \State $LSR_{ace} \gets \text{build shortcuts} (LSR, \varepsilon)$ [Alg.~\ref{alg::connect}]
    \end{algorithmic}
\Return $LSR_{ace}$
\end{algorithmic}
\label{alg::integration}
\end{algorithm}

%% file: includes/experiments.tex
\section{Simulation results}\label{sec:exp}

To validate the proposed approach, we consider a simulated box stacking task, shown in Fig. \ref{fig:sim_dissim} and referred to as \emph{hard} stacking task in \cite{bestlsr}. 
This setting allows to determine the true underlying state of each observation (exploited for evaluation purposes only) and therefore to automatically validate the effectiveness of each ACE component as well as of the entire ACE framework. 

The box stacking task is composed of a $3 \times 3$ grid where four boxes can be stacked on top of each other. The underlying state is defined by the geometrical arrangement of the boxes, where each box is considered unique. The action specifics $u$ is represented by the pick and place coordinates.  In each observation, we induce different lighting conditions as well as  $\approx 17\%$ random noise in the positioning of the boxes in each cell.  
 The following rules apply: \emph{i)} only one box can be moved at the time,  \emph{ii)} only one box can be placed in a single grid cell, \emph{iii)} boxes cannot float, and  \emph{iv)} a box can only be picked if no other box is on top of it.
Given the $3 \times 3$ grid and the above rules, the system exhibits exactly $288$ possible underlying states, and $|\mathcal{U}| = 48$ possible actions.

\subsection{Evaluation Criteria and Implementation Details}\label{sec:sim-scoring} 

To evaluate the effectiveness of the ACE paradigm in scarce data settings, we randomly sub-sample $80\%$, $75\%$, $60\%$, $50\%$, $40\%$, and $30\%$ of the original dataset $\mathcal{T}_{100}$ \cite{bestlsr} consisting of $2500$ pairs. We denote these sub-sampled dataset as $\mathcal{T}_{80}$, $\mathcal{T}_{75}$, $\mathcal{T}_{60}$, $\mathcal{T}_{50}$,  $\mathcal{T}_{40}$, and $\mathcal{T}_{30}$, respectively.  
We compare the combined ACE-LSR with the $\varepsilon$-LSR in \cite{iros} as well as with the ablated versions of each component of ACE, namely A-LSR for the augmentation step, C-LSR for connection step, and E-LSR for the exploration step. We additionally implement: \emph{i)} a baseline augmentation step, referred to as A$_b$-LSR, which generates similar pairs using the closest states in the latent space, and \emph{ii)} a baseline exploration step, referred to as E$_b$-LSR, which is a random explorer selecting a random action from the set $\mathcal{U}$ of the system actions and trying to apply it to the given state observation. 
  For the sake of space,  we omit the comparison of ACE-LSR  to other existing methods, but this can be found for the general LSR framework  in~\cite{bestlsr}.

We score all frameworks by the planning performance on $1000$ \emph{novel} start and goal states randomly selected from a holdout dataset composed of $2500$ observations.
We report the percentage of correct transitions, and the percentage of cases where all plans are correct, and where at least one of the suggested plan is correct, denoted by \textit{\% trans.}, \textit{\% all}, and \textit{\% any}, respectively.   
Furthermore,  to evaluate the augmentation component,  we report the number of newly identified similar pairs, \textit{\# pairs},  and the percentage of correct pairs among them, \textit{\% pairs},  using the ground truth underlying states. 
The connection component is similarly scored by measuring the number of new edges built in the graph, \textit{\# edges}, as well as the percentage of correct edges, \textit{\% edges}. Finally, the exploration component is evaluated by performing $n_e = 500$ exploration steps from random initial states and defining the percentage of valid exploratory actions, \textit{\% explore}.  For each score, we report mean and variance obtained with three different seeds for the MM model training.

The VAE modelling the MM is trained as in \cite{bestlsr} with latent dimension $16$. A DBSCAN-based~\cite{ester1996density} 
clustering algorithm is used for building the LSR. The hyperparameter~$\varepsilon$ is set to $\varepsilon = \mu_0  + w_\epsilon\cdot \sigma_0$ as in \cite{iros} where $\mu_0$ and $\sigma_0$ are the mean and  standard deviation  of the $L_1$ distances $\|z_1 - z_2\|_1$ among similar latent pairs $(z_1, z_2, a = 0)$, respectively, and $w_\epsilon$ is a scaling parameter.  We perform a grid search for $w_\epsilon$ in the interval $[-0.65, -0.05]$ with step size $0.1$. 
The LPM is a two-layer, 100 nodes MLP, while the Siamese network for the SM is a shallow two convolutional layer network with a latent space dimension $12$ as in \cite{ccpaper}. We train the Siamese network for $100$ epochs and  perform HDBSCAN~\cite{mcinnes2017hdbscan} clustering in the latent space $\mathcal{Z}'$ of the model. 
We set  search radius $r = \mu_0$ in Algorithm~\ref{alg::augment} since similar states should generally fall at a distance equal to the mean of similar states in the training dataset. While performing exploration, we additionally remove  the action obtained by reversing the last action from the set of possible actions 
and apply a reset of the system state each time an invalid action is undertaken.  

\subsection{Evaluation Results}

\begin{table*}[t!]
\resizebox{\textwidth}{!}{
\centering
\begin{tabular}{|l|l|l|l|l|l|l|l|l|}
\hline
Framework & \textit{\# pairs} &  \textit{\% pairs} &  \textit{\# edges} &  \textit{\% edges} &  \textit{\% explore} &  \textit{\% trans.} &  \textit{\% all} &  \textit{\% any} \\ \hline
$\varepsilon$-LSR \cite{iros} &   $-$    &  $-$     &    $-$   &   $-$    &    $-$     &  $60.5 \pm 4.4$ & $49.1 \pm 5.9$ & $49.6 \pm 5.7$   \\ \hline
A$_b$-LSR   &   \boldmath$1654 \pm 9.0$    &    $95.18$   &   $-$    &   $-$    &    $-$     &  $58.8 \pm 5.7$ & $32.3 \pm 19.2$ & $33.1 \pm 19.2$    \\ \hline
A-LSR    &    $15 \pm 11.0$   & \boldmath$100$  &    $-$   &    $-$   &     $-$    &  $64.0 \pm 13.5$ & $56.5 \pm 12.1$ & $56.8 \pm 12.3$  \\ \hline
C-LSR    &    $-$   &     $-$  &   \boldmath$479.2 \pm 52.1$ & $95.1 \pm 0.4$ & $-$ & $87.7 \pm 0.6$ & $57.1 \pm 9.4$ & $63.0 \pm 7.9$
 \\ \hline
E$_b$-LSR   &    $-$     &    $-$   &   $-$    &   $-$    &     $7.8 \pm 0.0$ & $65.9 \pm 0.2$ & $51.4 \pm 5.7$ & $52.3 \pm 5.3$
   \\ \hline
E-LSR    &  $-$  &    $-$   &    $-$   &     $-$  &    \boldmath$97.7 \pm 0.4$ & $80.7 \pm 2.0$ & $62.2 \pm 3.0$ & $65.7 \pm 3.1$
\\ \hline
ACE-LSR      &       $15 \pm 11.0$    & \boldmath$100$      &  $401.0 \pm 15.3$ & \boldmath$96.6 \pm 1.0$ & $97.1 \pm 1.6$ & \boldmath$93.1 \pm 1.9$ & \boldmath$79.2 \pm 4.2$ & \boldmath$82.6 \pm 4.0$
     \\ \hline
\end{tabular}
}
\vspace{-5pt}
\caption{Evaluation results obtained on the box stacking task  with  $\mathcal{T}_{50}$ using $\varepsilon$-LSR as well as its combination with the baseline augmentation and exploration methods, the individual components of ACE paradigm and all the ACE components.
The symbol $-$ denotes that the respective score is not relevant to the framework. See Sec.~\ref{sec:sim-scoring} for details. Best results in bold. }
\label{tab::resultsall}
\vspace{-5pt}
\end{table*}

\begin{figure}
    \centering
    \includegraphics[width=\linewidth]{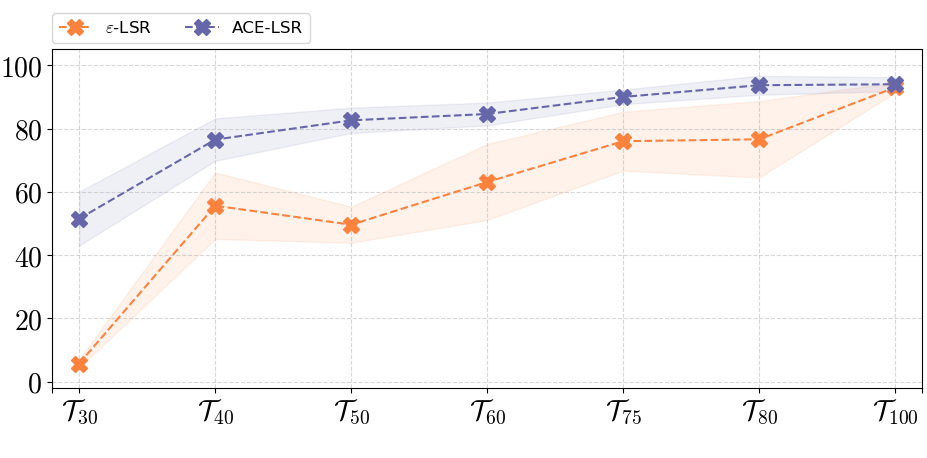}
    \vspace{-25pt}
    \caption{ Planning performance on the box stacking task in terms of \emph{\% any} using $\varepsilon$-LSR \cite{iros} (orange) and ACE-LSR (blue) trained on the subsampled datasets.
}
    \label{fig:results_data}
\end{figure}

Figure~\ref{fig:results_data} shows the planning performance in terms of \mbox{\textit{\% any}} score across the considered  subsampled datasets when the proposed ACE paradigm is used (blue) and not (orange). Cross marks denote mean values, while the transparency represents the variance.  We can observe that ACE-LSR boosts the planning performance compared to the $\varepsilon$-LSR~\cite{iros} for each subsampled dataset and is particularly essential  in case of very scarce datasets, e.g. $\mathcal{T}_{30}$-$\mathcal{T}_{50}$. For example,  average improvements equal to $\approx 45,\,21,\,33\%$ are observed for $\mathcal{T}_{30}$,$\mathcal{T}_{40}$,$\mathcal{T}_{50}$, reaching $\approx 51.5, 76.5, 82.6\%$, respectively. 
Obviously, the improvement is much more significant with scarce datasets, while the performance is almost saturated with $\mathcal{T}_{100}$, reaching  $93\%$ and $94\%$ with $\varepsilon$-LSR and ACE-LSR, respectively.
%
    

Table \ref{tab::resultsall} shows the results of the ablation study for the components of the ACE paradigm compared with the $\varepsilon$-LSR. 
We report the complete scoring described in Sec.~\ref{sec:sim-scoring} obtained using $\mathcal{T}_{50}$ which consists of half the data used in \cite{iros}. 
We observe that data scarcity leads to \textit{unsatisfactory} planning performance of the $\varepsilon$-LSR, reaching only average {\textit{\% any}} score of $49.6\%$ with {\textit{\% trans}} equal to $60.5\%$.  
No improvement but rather a decrease of performance is recorded with the baseline augmentation step, i.e., with A$_b$-LSR (row~2). This builds 
 $1654$ new similar pairs among which $\approx95\%$ are correct. However, these new pairs deteriorate the structure of the latent space, resulting in {\textit{\% any}} equal to $33.1 \%$ only with a decrease of $\approx 16\%$. This suggests that simply adding new correct pairs does not necessarily induce improved performance if they are not carefully selected. 
In contrast, our augmentation algorithm A-LSR (row~3) produces only  $15$ new similar pairs on average that are $100\%$ correct, thus boosting the planning performance in terms of \textit{\% any} to average $56.8 \%$. 
Our connection algorithm in C-LSR (row~4). builds $\approx479$ new shortcuts that are $\approx 95\%$ correct. These yield to much higher planning scores, reaching  average $87.7\%$ and $63\%$ for \textit{\% trans.} and \textit{\% any}, respectively. 
Concerning the exploration phase, only $7.8\%$ of the moves (\textit{\% explore} score) attempted by the baseline random explorer in E$_b$-LSR (row~5) are correct. 
This results in $\approx 2.5$ percentage point enhancement of the planning performance in terms of \textit{\% any} compared to $\varepsilon$-LSR. 
On the other hand, a substantial improvement in planning performance is recorded when employing our exploration algorithm E-LSR (row~6). More specifically, $\approx 97.7\%$ of the $n_e=500$ exploration moves are found to be valid,  resulting in average $80.7\%$ and $65.7\%$ for \textit{\% trans.} and \textit{\% any} scores, respectively. This result suggests the effectiveness of the proposed SM models for proposing exploration actions, which are found almost always to be correct. Examples of suggested actions by the SM for the stacking task are reported in Fig.~\ref{fig:sm_master}-left. 
Finally, the combined ACE-LSR approach (row~7) significantly outperforms all of the above mentioned frameworks, leading to an improvement $>30\%$ in terms of \textit{\% trans.,\% all,\% any} compared to the $\varepsilon$-LSR and  reaching a final \textit{\% any} performance of $82.6\%$.


\section{Experimental Results} 
\begin{figure}
    \centering
    \includegraphics[width=\linewidth]{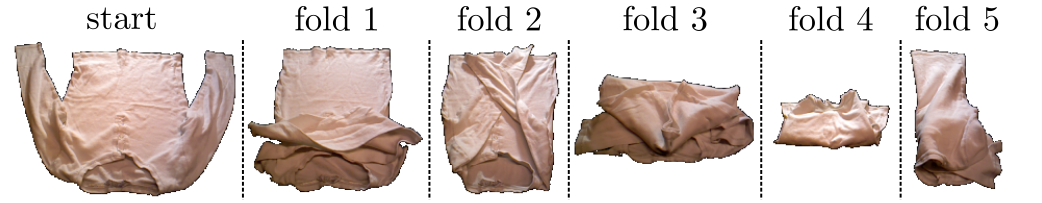}
    \caption{Start and five goal configurations of the T-shirt.}
    \label{fig:folds}
\end{figure}

To further validate the effectiveness of the  ACE paradigm, we perform a real world T-shirt folding task as in \cite{iros}.
In this task, the goal is to generate and execute visual action plans from a start configuration to five different goal configurations, shown in Fig. \ref{fig:folds}. 

Let 
$\mathcal{F}_{100}$ be the training dataset used in \cite{iros} containing a total of $1150$ pairs, and let $\mathcal{F}_{50} \subset \mathcal{F}_{100}$ be a scarce dataset consisting of $50\%$ randomly subsampled pairs. We use the same set of parameters and architectures as in Sec.~\ref{sec:sim-scoring} unless otherwise specified. Since the action specifics $u$ is composed of \emph{pixel} position pick and place coordinates, the action space is much larger as in the simulation task. In order for the SM to be able to suggest meaningful actions, we discretize the action space into bins and use the mean actions of each bin. We choose a bin size of $\approx 15\%$ of the image space which results in $107$ unique actions. 
Furthermore, we group the observations only based on the similarity of the pick action as this enables more flexible exploration. Examples of suggested actions for the T-shirt folding task are shown in Fig.~\ref{fig:sm_master}-right. 
The SM, trained for $200$ epochs, is then used in the augmentation step  obtaining $13$ new similar pairs (Algorithm~\ref{alg::augment}). 
When applying the connection algorithm using the SM and LPM-R we obtain $55$ novel edges in the graph. In order to obtain more novel connections, we increase the LPM-R reliability check by a factor of $1.5$ since both the scarcity and diversity of the actions make a reliable prediction more  challenging.
Lastly, we execute Algorithm~\ref{alg::explore} for $n_e = 20$ exploration steps. Note that exploration in the folding task is much less constrained, and therefore some explorations can lead to completely novel folding sequences not observed in the collected training dataset $\mathcal{T}_o$.
Including the newly obtained action pairs yields the final ACE-LSR (built with $w_\varepsilon=1$) that we compare with $\varepsilon$-LSR ($w_\varepsilon=1.4$)  trained on $\mathcal{F}_{50}$ in Tab.~\ref{tab::resultsall}. We repeat each fold five times and report the number of successful trials when the entire fold is performed successfully. The execution videos, visual action plans as well as the exploration can be seen on the project website\footnote{\url{https://visual-action-planning.github.io/ace/}}.  Moreover, an example of a generated visual action plan is provided in Fig.~\ref{fig:overview}. 

The ACE-LSR outperforms the $\varepsilon$-LSR in all folds except for fold $4$, and reaches a total system success rate over all five folds of $80\%$, matching the performance reported in \cite{iros} using only half the training data. We observe that the $\varepsilon$-LSR does not have enough data to distinguish fold $1$ from fold $2$ as it always performs fold $2$ regardless of the fold goal state. On the contrary, ACE-LSR is able to successfully distinguish them and execute the correct fold most of the times. Furthermore, the $\varepsilon$-LSR is not able to reliably execute fold $5$ as it is missing the final step to complete it, while ACE-LSR is able to perform it in $4/5$ cases.

\begin{table}[]
\begin{tabular}{|l|l|l|l|l|l|}
\hline
Framework & fold 1 & fold 2 & fold 3 & fold 4 & fold 5 \\ \hline
$\varepsilon$-LSR   &   $0/5$     &     \boldmath$5/5$   &   $0/5$     &  \boldmath$5/5$   &  $1/5$      \\ \hline
ACE-LSR   &    \boldmath$4/5$    &   \boldmath$5/5$     &    \boldmath$3/5$    &   $4/5$     &     \boldmath$4/5$   \\ \hline
\end{tabular}
\vspace{-5pt}
\caption{System performance results on the T-shirt folding task with $\mathcal{F}_{50}$ for $\varepsilon$-LSR and ACE-LSR on five different folds, each repeated five times. Best results in bold.  }
\vspace{-2pt}
\end{table}




%% file: includes/conclusions.tex
\section{Conclusions}
In this work, we presented the ACE paradigm that addresses data scarcity problem for visual action planning. We built upon the Latent Space Roadmap framework and 
introduced \emph{i)} a novel Suggestion Model (SM), that given an observation, suggests possible actions in that state, and \emph{ii)} a Latent Prediction Model (LPM) that, given a latent state and an action, predicts the next  latent state. Combining these modules, we \textbf{A}ugmented the dataset to identify new similar pairs for training, identified new valid  edges in the LSR to increase its \textbf{C}onnectivity, and \textbf{E}xplored the latent space efficiently to reach potential undiscovered states. We validated the ACE paradigm on a simulated box stacking task and a real-world T-shirt folding task on several levels of data scarcity. As future work, we aim to extend this paradigm to different contexts, such as RL. 
